\documentclass[11pt]{article}
\usepackage[final]{acl}
\usepackage{times}
\usepackage{latexsym}
\usepackage[T1]{fontenc}
\usepackage[utf8]{inputenc}
\usepackage{microtype}
\usepackage{inconsolata}
\usepackage{graphicx}
\usepackage{booktabs}
\usepackage{amsmath,amssymb}

\newcommand{\rhot}{\rho_{\text{tr}}}
\newcommand{\dec}{\textsc{Dec}}
\newcommand{\cfn}{\textsc{CF}_{N}}
\newcommand{\cft}{\textsc{CF}_{T}}
\newcommand{\conf}{\textsc{Conflict}}
\newcommand{\leak}{\textsc{Leak}}

\newcommand{\absCFcebabQFive}{.25}

\newcommand{\absamznegFive}{.07--.09}
\newcommand{\absamznegNine}{.07--.09}
\newcommand{\amzFpNeg}{.66}
\newcommand{\amzFpPos}{.68}
\newcommand{\amzNegNeg}{.64}
\newcommand{\amzNegPos}{.30}
\newcommand{\cfNnbMax}{.89}
\newcommand{\cfNnbMin}{$-$.10}
\newcommand{\cfTmax}{1.19}
\newcommand{\cfTmin}{.84}
\newcommand{\civilAccFwd}{13\%}
\newcommand{\civilAccRev}{29\%}
\newcommand{\corrCFmax}{+.38}
\newcommand{\corrCFmin}{$-$.61}

\newcommand{\cosMaxcebab}{.83}

\newcommand{\cosMincebab}{.65}

\newcommand{\dirNegamznegFive}{$-$.07--$-$.06}

\newcommand{\eqCFmaxamzfpGNine}{.25}

\newcommand{\eqCFmaxamzfpQNine}{.22}

\newcommand{\eqCFmaxbiosQNine}{.42}

\newcommand{\eqCFmaxcebabGNine}{.63}

\newcommand{\eqCFmaxcebabQNine}{.83}

\newcommand{\eqCFminamzfpGNine}{.06}

\newcommand{\eqCFminamzfpQNine}{.11}

\newcommand{\eqCFminbiosQNine}{.09}

\newcommand{\eqCFmincebabGNine}{.25}

\newcommand{\eqCFmincebabQNine}{.54}

\newcommand{\eqNbiosQNine}{65}

\newcommand{\fpCFlramzfpQNine}{.17}

\newcommand{\fpCFlrbiosQNine}{.22}

\newcommand{\fpCFneamzfpQNine}{.07}

\newcommand{\fpCFnebiosQNine}{.07}

\newcommand{\fpHibiosQNine}{+.002}

\newcommand{\fpLobiosQNine}{+.000}

\newcommand{\frCebabAll}{.41--1.15}
\newcommand{\frCebabBoth}{.26--.89}

\newcommand{\frSelFullMax}{.39}
\newcommand{\frSelFullMin}{.04}
\newcommand{\frSpearMin}{.87}

\newcommand{\geomMAEcebab}{.11}

\newcommand{\gptPlaceboAccMax}{.97}
\newcommand{\gptPlaceboAccMin}{.88}
\newcommand{\incamzfp}{.11--.16}
\newcommand{\incamzneg}{.09--.11}
\newcommand{\jvBadRate}{.17}
\newcommand{\jvGoodRate}{.83}
\newcommand{\jvGptQ}{.92}
\newcommand{\jvHumanQ}{.90--.91}
\newcommand{\jvN}{529}
\newcommand{\jvNbad}{89}
\newcommand{\jvPrec}{.94}
\newcommand{\jvQwenQ}{.77--.84}
\newcommand{\jvRecall}{.50}
\newcommand{\jvRejBad}{.84}
\newcommand{\kappaMax}{.55}
\newcommand{\kappaMin}{.06}

\newcommand{\nbCFmaxamznegQNine}{.12}

\newcommand{\nbCFmaxcebabQFive}{.28}

\newcommand{\nbCFmaxcivilGNine}{.50}

\newcommand{\nbCFmaxcivilQFive}{.08}
\newcommand{\nbCFmaxcivilQNine}{.45}

\newcommand{\nbCFminamznegQNine}{.03}

\newcommand{\nbCFmincebabQFive}{.02}

\newcommand{\nbCFmincivilGNine}{.30}

\newcommand{\nbCFmincivilQFive}{.04}
\newcommand{\nbCFmincivilQNine}{.24}

\newcommand{\plBadMax}{.10}
\newcommand{\plCleanMax}{.94}
\newcommand{\plCleanMin}{.86}
\newcommand{\plFactMax}{.88}
\newcommand{\plFactMin}{.52}

\newcommand{\plMixBoth}{.18}
\newcommand{\plStrBothMax}{.18}
\newcommand{\plStrGptMax}{.68}
\newcommand{\plStrGptMin}{.58}
\newcommand{\placeboMax}{.06}
\newcommand{\placeboMin}{.01}

\newcommand{\placebocebabQFive}{.06}

\newcommand{\plcamznegNine}{.03}

\newcommand{\qwenPlaceboAccMax}{.72}
\newcommand{\qwenPlaceboAccMin}{.44}
\newcommand{\rdAgree}{.73}
\newcommand{\rdKappa}{.45}
\newcommand{\rdNotOkFp}{.18--.32}
\newcommand{\rdNotOkNeg}{.36--.42}
\newcommand{\rdWrongFp}{.02--.14}
\newcommand{\rdWrongNeg}{.12--.22}
\newcommand{\rdWrongNegAdd}{.24--.28}

\newcommand{\robAllValMin}{.77}
\newcommand{\robDirMinNine}{.93}
\newcommand{\robSDMin}{.87}

\newcommand{\selFullConfCFmax}{.34}
\newcommand{\selFullConfCFmin}{.04}
\newcommand{\selFullConfDecMeanWorst}{.007}

\newcommand{\selSensConfCFmax}{.15}
\newcommand{\selSensConfCFmin}{.02}
\newcommand{\selSensConfDecMeanWorst}{.012}
\newcommand{\selSensConfDecWorst}{.020}

\newcommand{\sgnCFcebabQFive}{.07}

\newcommand{\sgnamzfpNine}{.12--.17}

\newcommand{\sgnamznegNine}{.05--.06}
\newcommand{\synCFRangeStdMax}{.87}
\newcommand{\synCFRangeStdMin}{.83}
\newcommand{\synDecRangeNe}{.09}
\newcommand{\synDecRangeStd}{.03}

\title{Decodability Is Not Validity: Evaluating Nuisance Sensitivity\\in Language Model Concept Probes}

\author{Ayal Klein \\
  \texttt{ayalk@ariel.ac.il} \\
  \small\url{https://github.com/kleinay/decodability-is-not-validity}}

\begin{document}
\maketitle

\begin{abstract}
Two concept probes can be equally accurate at predicting a concept from a language model's hidden states while their scores contain different mixtures of the target concept and a correlated nuisance attribute. In four target--nuisance settings (two aspect sentiments in CEBaB; occupation and gender in BiasBios; Amazon review sentiment with negation, and with first-person framing), two LLMs, and a population of concept extractors, we compare decodability with discriminant evidence: the score's dependence on the nuisance at fixed target, and its response to input edits intended to change only the nuisance, filtered by two LLM judges and compared against placebo edits. Under correlated training data, extractors that are statistically equivalent in held-out AUROC differ up to several-fold in nuisance sensitivity in three of the four settings, and decodability is not a reliable proxy for it: their rank correlation changes sign across settings. The effect follows the probe's training data, not the nuisance's natural association with the target: first-person framing, unrelated to sentiment in natural reviews, is absorbed at least as much as negation, which co-occurs with negative sentiment, and decorrelated training removes the systematic effect. Selecting extractors with nuisance-aware development diagnostics, which need nuisance labels but no counterfactuals, lowers counterfactual nuisance sensitivity wherever the default choice is sensitive, at a small accuracy cost. Concept measurements should be validated, not just decoded.
\end{abstract}

\section{Introduction}

Probes, concept activation vectors and subspace readouts turn a hidden state into a scalar proxy for a concept $T$ in the input \citep{alain2017understanding,kim2018tcav,belinkov2022probing,zou2023representation,marks2024geometry}; the usual evidence for that reading is decodability, i.e.\ predicting held-out labels of $T$. Decodability cannot identify \emph{what} the score measures. Call a \emph{nuisance} $N$ an attribute that is distinct from $T$ but may be correlated with it (e.g., food sentiment when $T$ is service sentiment). Take binary $T$ and $N$ and two scores, $s_1=T$ and $s_2=T+\alpha N$ with $0<\alpha<1$. Every input with $T{=}1$ scores at least 1 under $s_2$ and every input with $T{=}0$ at most $\alpha<1$, so $s_2$, like $s_1$, ranks all positives above all negatives: both have AUROC 1. Yet only $s_2$ changes when $N$ changes and $T$ does not.

In measurement theory this is a question of construct validity, an evidential, multidimensional property of a score's interpretation \citep{cronbach1955construct,messick1995validity,jacobs2021measurement,bean2025measuring}. Part of that evidence is \emph{discriminant} \citep{campbell1959convergent}: measures of correlated constructs may correlate, but the score should not depend on the other construct \emph{once the target is held fixed}.

That accuracy cannot detect such dependence is not new: probes trained on spuriously correlated data can rely on the spurious attribute \citep{kumar2022probing}, probing accuracy need not reflect use by the model \citep{hewitt2019designing,ravichander2021probing,elazar2021amnesic,tiwari2026decodability,sudheendra2026decodability}, and interventions on representations trade completeness against selectivity \citep{canby2025reliable,huang2024ravel}. Recent work tests whether probes and SAEs disentangle known concepts \citep{mueller2026isolation}, compares concept-subspace estimators \citep{naowarat2026framework}, and shows that accurate probes can be misaligned \citep{lysnaeslarsen2025probing}, that probe readings depend on design choices \citep{noel2026probe} and correlated features \citep{shen2026beyond}, and that embedding-based measures need counterfactual validation \citep{li2026proxy}. We ask a narrower question about scalar readouts: among accuracy-matched extractors, how much does sensitivity to a named nuisance under input edits vary, does it persist without training-set correlation, and can development diagnostics choose better? We answer with a population study over four settings and two models, using judge-filtered, placebo-controlled edits, including two nuisances edited on the same reviews (\S\ref{sec:results}), and compare selection criteria (\S\ref{sec:selection}). Our interventions edit the \emph{input text} of frozen models: unlike work asking whether a decodable direction drives behavior \citep{tiwari2026decodability,sudheendra2026decodability}, we ask what a score tracks in the input.

\section{Measuring nuisance selectivity}
\label{sec:battery}

An \emph{extractor} maps a hidden state $h$ to $s(h)=w^\top z(h)+b$, with $z$ the standardization on training data; subspace methods have a linear readout too, so every extractor is a scalar score. Each setting supplies a binary target $T$, a binary nuisance $N$, and edits of the input. Let \mbox{$\rho=P(N{=}1\mid T{=}1)$} \mbox{${}=P(N{=}0\mid T{=}0)$} with balanced $T$. Extractors are trained at $\rhot\in\{0.5,0.7,0.9\}$; test metrics reweight the four $(T,N)$ cells of one fixed test pool.

\noindent\textbf{\dec{}}: AUROC of $s$ for $T$ at $\rho=\rhot$ (in-distribution decodability, the standard criterion).\\
\textbf{$|$\leak{}$|$}: $\tfrac12\sum_{t}|\mathrm{AUROC}(s\to N\mid T{=}t)-0.5|$, dependence on $N$ with the target held fixed.\\
\textbf{\conf{}}: AUROC for $T$ on the cells with $T\neq N$ (a fully reversed correlation). A stress test, not a validity measure: there, $s=-N$ also ranks $T$ perfectly.\\
\textbf{$\cfn$}: mean change in $s$ under edits that flip only $N$, signed toward the $N$ value that co-occurs with $T{=}1$ in training, divided by the mean score gap between the $T$ classes on the decorrelated pool ($\Delta_T$). As a signed mean can hide opposite changes, we also report $|\cfn|$ (mean absolute change over $\Delta_T$) and compare with \emph{placebo} edits that change neither variable.\\
\textbf{$\cft$}: the same quantity for edits that flip $T$ (CEBaB only): convergent evidence.

\section{Setup}
\label{sec:setup}

\begin{table}[t]
\centering\small\setlength{\tabcolsep}{2pt}
\resizebox{\columnwidth}{!}{\begin{tabular}{@{}llccc@{}}
\toprule
 & target $T$ / nuisance $N$ & $N$ edits & judged by both & accepted \\
\midrule
CEBaB & service / food sentiment & 210 & 210 & 41\% \\
BiasBios & surgeon vs.\ nurse / gender & 1561 & 1561 & 93\% \\
Amazon-Neg & negative sentiment / negation & 1427 & 1427 & 58\% \\
Amazon-1P & negative sentiment / first person & 1083 & 1083 & 68\% \\
Civil$^*$ & toxic / Muslim or LGBTQ & 762+400 & 762+400 & 13\%, 29\% \\
\bottomrule
\end{tabular}
}
\caption{Settings. Edits: generated nuisance-edit pairs (Amazon: those passing automatic checks); judged by both: pairs rated by both LLM judges; accepted: share of jointly judged pairs both accept (\S\ref{sec:setup}). $^*$Civil (rule-based identity swaps, forward/reverse) is reported in Appendix~\ref{app:civil} only.}
\label{tab:data}
\end{table}

\paragraph{Data and edits} (Table~\ref{tab:data}). \textbf{CEBaB} \citep{abraham2022cebab}: restaurant reviews with aspect labels and human-written counterfactual edits; we pair each original test review with edits that flip food (or, for $\cft$, service) sentiment and leave the other aspect's label unchanged; splits are disjoint by review family. \textbf{BiasBios} \citep{dearteaga2019bias}: surgeon vs.\ nurse biographies; $N$ is the subject's gender, edited in both directions by an LLM (gpt-oss-120b) that swaps the subject's pronouns, gendered nouns and first name and nothing else (automatically checked). \textbf{Amazon} \citep{mcauley2013hidden,zhang2015character}: short product reviews (8--40 words), $T$ = negative sentiment, with two nuisances defined by word lists, so whether an edit flips them is checked exactly: \emph{negation} (\emph{not}, \emph{n't}, \emph{never}, \emph{no}, \dots; in training data present in \amzNegNeg{} of negative and \amzNegPos{} of positive reviews) and \emph{first person} (the writer refers to themselves: \emph{I}, \emph{my}, \emph{we}, \dots; \amzFpNeg{} vs.\ \amzFpPos{}, i.e.\ uncorrelated). Each is its own setting (Amazon-Neg, Amazon-1P). gpt-oss-120b proposes phrase replacements that add or remove the nuisance while keeping the meaning and the strength of the opinion (``doesn't work''$\to$``fails''; ``Would recommend''$\to$``I would recommend''), applied by exact string match (Appendix~\ref{app:edits}). We keep up to 1{,}100/250/400 training/development/test texts per $(T,N)$ cell and train on samples of 2{,}000.

\paragraph{Edit validation.} Two judge models, neither identical to a probed model (Qwen3-32B, run locally, and gpt-oss-120b; for Amazon, Qwen3.8-27B and gpt-oss-120b via an API; the Qwen judges share a model family with Qwen2.5-7B), see each pair in random order, without being told which text is edited or which attribute should change, and say whether the target and the nuisance are the same, rate each text's fluency, and flag other meaning changes. Checked against human labels and planted errors (Appendix~\ref{app:judgeval}), a nuisance edit is \emph{accepted} if both judges find the target unchanged and the nuisance changed, with no major other meaning change and no fluency drop. For Amazon the nuisance flip is checked by the word list instead (the judges missed real flips in a pilot), and both judges must also find the strength of the opinion unchanged. All counterfactual results use accepted pairs only (Appendix~\ref{app:edits}). Placebo edits replace neutral words with synonyms.

\paragraph{Models and extractors.} Qwen2.5-7B-Instruct \citep{qwen2025qwen25} and Gemma-2-9B-it \citep{gemma2024gemma2}, frozen; residual stream mean-pooled over non-special tokens at 7 depths. Per layer: difference of class means; $\ell_2$ logistic regression ($C\in\{10^{-3},\dots,10\}$); partial least squares ($k\in\{2,4,8,16\}$; $k{=}1$ equals the mean difference and is dropped); and a \emph{nuisance-erased} probe that projects out the within-class nuisance mean differences, estimated on the training sample only (cf.\ \citealp{ravfogel2020null,belrose2023leace}) before logistic regression. The first three families use no nuisance labels (\emph{standard}, 70 per setting); the erased probe adds 35. Each is fit on three training samples. An extractor is \emph{near-best} if its test \dec{} is within 0.01 of the best, and \emph{equivalent} to the best if a paired-bootstrap 90\% CI of the \dec{} difference lies inside $\pm0.01$ (TOST at $\alpha=0.05$). Because the reference is selected on the test pool, these equivalence results are descriptive; the selection comparisons (\S\ref{sec:selection}) choose on development data only.

\section{Results}
\label{sec:results}

\begin{figure*}[t]
\centering
\includegraphics[width=\textwidth]{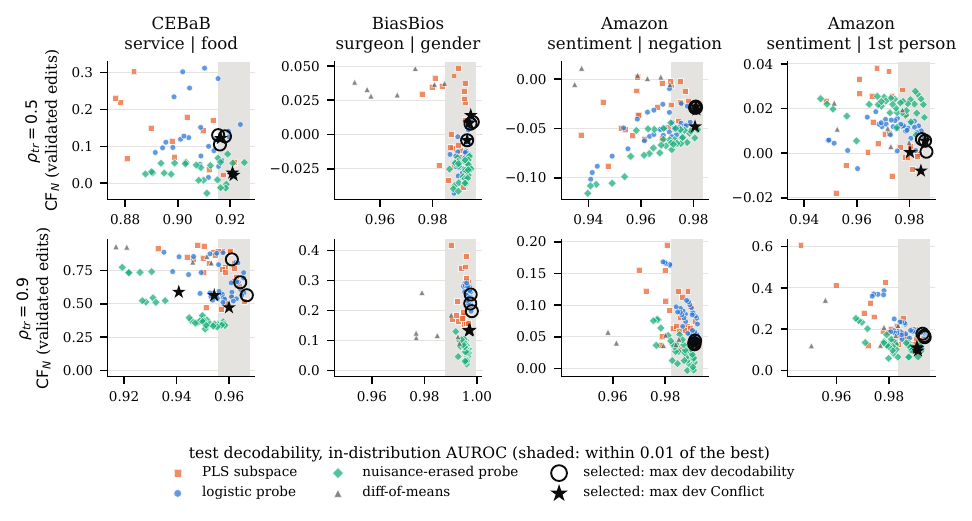}
\caption{Competitive extractors (\dec{} within 0.05 of the best; Qwen2.5-7B; mean over 3 training samples): test decodability vs.\ nuisance sensitivity $\cfn$ on accepted edits (0: no net shift; 1: the nuisance edit shifts the score as much as a change of target class; note the different $y$ scales). Black symbols: the extractor that each training sample selects on development data by maximal \dec{} (ring) or maximal \conf{} (star), standard extractors only. Gemma-2-9B: Appendix~\ref{app:more}.}
\label{fig:main}
\end{figure*}

\begin{table}[t]
\centering\small\setlength{\tabcolsep}{2pt}
\resizebox{\columnwidth}{!}{\begin{tabular}{@{}llccccc@{}}
\toprule
 & & best & \#near & $\cfn$ near-best & $\cfn$ equiv. & \conf{} \\
 & & \dec{} & (equiv.) & range & range & range \\
\midrule
\multicolumn{7}{@{}l}{\emph{$\rhot=0.9$ (correlated training data)}}\\
CEBaB & Q & .966 & 37 (9) & .46--.89 & .54--.83 & .44--.69 \\
 & G & .979 & 39 (8) & .26--.72 & .25--.63 & .57--.82 \\
BiasBios & Q & .998 & 66 (65) & .09--.42 & .09--.42 & .83--.97 \\
 & G & .998 & 66 (61) & .09--.33 & .11--.33 & .92--.97 \\
Amazon-Neg & Q & .992 & 51 (45) & .03--.12 & .03--.12 & .87--.95 \\
 & G & .994 & 64 (60) & .02--.12 & .02--.11 & .87--.96 \\
Amazon-1P & Q & .993 & 40 (27) & .11--.25 & .11--.22 & .85--.95 \\
 & G & .994 & 62 (49) & .05--.31 & .06--.25 & .84--.96 \\
\bottomrule
\end{tabular}
}
\caption{Standard extractors that are near-best (test \dec{} within 0.01 of the best) and, in parentheses, statistically \emph{equivalent} to the best (TOST, margin 0.01). Q: Qwen2.5-7B; G: Gemma-2-9B. $\rhot\in\{0.5,0.7\}$ and Civil: Appendix~\ref{app:more},~\ref{app:civil}.}
\label{tab:near}
\end{table}

\paragraph{Accuracy-matched extractors differ in nuisance sensitivity.} With correlated training data ($\rhot=0.9$), extractors equivalent to the best in decodability span $\cfn$ from \eqCFmincebabQNine{} to \eqCFmaxcebabQNine{} on CEBaB (Qwen; Gemma \eqCFmincebabGNine--\eqCFmaxcebabGNine): for some, rewriting only the food sentence of a review shifts the ``service'' score most of the way to the other service class. On BiasBios \dec{} is at ceiling for nearly all extractors (\eqNbiosQNine{} equivalent to the best), while gender edits shift the occupation score by \eqCFminbiosQNine--\eqCFmaxbiosQNine{} of the surgeon--nurse gap; on Amazon-1P, adding or removing a first-person reference shifts the sentiment score of equivalent extractors by \eqCFminamzfpQNine--\eqCFmaxamzfpQNine{} (Gemma \eqCFminamzfpGNine--\eqCFmaxamzfpGNine). On Amazon-Neg, every near-best extractor is only weakly sensitive (\nbCFminamznegQNine--\nbCFmaxamznegQNine). Rankings by $\cfn$ agree on all vs.\ accepted edits ($\geq$\robAllValMin), under SD normalization ($\geq$\robSDMin) and, at $\rhot=0.9$, between edit directions ($\geq$\robDirMinNine; Appendix~\ref{app:edits}); placebo edits move scores little (median $|\cfn|$ \placeboMin--\placeboMax).

\paragraph{Same reviews, two nuisances.} First person is unrelated to sentiment in natural reviews, yet under the same training correlation ($\rhot=0.9$) the median $\cfn$ of competitive extractors is \sgnamzfpNine{} for first person and \sgnamznegNine{} for negation (an increase over decorrelated training of \incamzfp{} vs.\ \incamzneg): the probe's training data, not the nuisance's natural association, produces the reliance. Negation edits move scores by a similar absolute amount at any training correlation (median $|\cfn|$ \absamznegFive{} at $\rhot=0.5$, \absamznegNine{} at 0.9; placebo \plcamznegNine), and at $\rhot=0.5$ against the nuisance direction, mostly when a negation is removed (\dirNegamznegFive): consistent with ``not good''$\to$``bad'' sharpening the opinion, which the strength check does not fully exclude, rather than with reliance on the nuisance.

\paragraph{With decorrelated training, the directional effect mostly disappears.} At $\rhot=0.5$, near-best extractors show little net nuisance shift on BiasBios and both Amazon settings (signed $\cfn$ within $\pm0.08$) and less on CEBaB (\nbCFmincebabQFive--\nbCFmaxcebabQFive, Qwen), where example-level shifts remain (median $|\cfn|$ \absCFcebabQFive{} vs.\ signed \sgnCFcebabQFive; placebo \placebocebabQFive, for smaller edits). Training-set correlation drives the systematic reliance on the nuisance. A second source is geometric: on CEBaB the target and within-class food directions are aligned (cosine \cosMincebab--\cosMaxcebab), and this observational geometry approximately predicts the food sensitivity of the difference of means trained on decorrelated data (mean absolute error \geomMAEcebab; synthetic illustration in Appendix~\ref{app:synthetic}).

\paragraph{Decodability is not a reliable proxy.} Among competitive standard extractors at $\rhot=0.9$, the rank correlation between \dec{} and $\cfn$ ranges from \corrCFmin{} to \corrCFmax{} and changes sign across settings (descriptive intervals: Appendix~\ref{app:more}). Target edits do not arbitrate either: on CEBaB every near-best extractor responds fully to \emph{service} edits ($\cft$ \cfTmin--\cfTmax) while its food sensitivity ranges over \cfNnbMin--\cfNnbMax.

\paragraph{A fixed-method comparison.} With layer and hyperparameter chosen on development data, the logistic and the nuisance-erased probe are equivalent in \dec{} at $\rhot=0.9$ on BiasBios (90\% CI [\fpLobiosQNine, \fpHibiosQNine], Qwen) and Amazon-1P, yet $\cfn$ is \fpCFlrbiosQNine{} vs.\ \fpCFnebiosQNine{} and \fpCFlramzfpQNine{} vs.\ \fpCFneamzfpQNine{} (Appendix~\ref{app:more}). Erasure lowers but does not remove nuisance sensitivity.

\subsection{Selecting for nuisance selectivity}
\label{sec:selection}

\begin{table}[t]
\centering\small\setlength{\tabcolsep}{2pt}
\resizebox{\columnwidth}{!}{\begin{tabular}{@{}llcccc@{}}
\toprule
 & & \multicolumn{3}{c}{standard extractors} & + erased \\
 & & max \dec{} & max \conf{} & min $|$\leak{}$|$ & max \conf{} \\
\midrule
CEBaB & Q & .69 & .54 ($-$.012) & .50 ($-$.013) & .34 ($-$.007) \\
 & G & .41 & .26 ($-$.005) & .27 ($-$.004) & .11 ($-$.004) \\
BiasBios & Q & .22 & .13 ($-$.001) & .09 ($-$.004) & .03 ($-$.002) \\
 & G & .24 & .14 ($-$.002) & .12 ($-$.006) & .02 ($-$.002) \\
Amazon-Neg & Q & .04 & .04$^\dagger$ (+.000) & .07$^\dagger$ ($-$.007) & $-$.00 ($-$.000) \\
 & G & .04 & .03$^\dagger$ (+.000) & .06$^\dagger$ ($-$.004) & $-$.00 ($-$.001) \\
Amazon-1P & Q & .17 & .11 ($-$.002) & .11 ($-$.002) & .07 ($-$.002) \\
 & G & .10 & .08$^\dagger$ ($-$.002) & .08$^\dagger$ ($-$.007) & .04 ($-$.003) \\
\bottomrule
\end{tabular}
}
\caption{Selecting an extractor on development data at $\rhot=0.9$: test $\cfn$ (accepted edits) of the extractor chosen by maximal \dec{} (standard practice), maximal \conf{}, or minimal $|$\leak{}$|$ among extractors within 0.01 of the best development \dec{}; change in test \dec{} in parentheses. Means over 3 training samples; $^\dagger$: $\cfn$ did not fall in every sample. More criteria and Civil: Appendix~\ref{app:more}.}
\label{tab:select}
\end{table}

Table~\ref{tab:select} compares the usual choice by development accuracy with nuisance-aware criteria that need nuisance labels on development data but no counterfactuals. Among standard extractors, choosing by \conf{} lowers $\cfn$ wherever the default choice is sensitive (CEBaB, BiasBios, Amazon-1P: by \selSensConfCFmin--\selSensConfCFmax, in every training sample but one) at a mean \dec{} cost of at most \selSensConfDecMeanWorst{} (\selSensConfDecWorst{} in the worst sample), and lowers $|\cfn|$ as well; on Amazon-Neg, where the default is already insensitive, it changes little. The leakage-constrained rule is less reliable (it raises $\cfn$ on Amazon-Neg). When the nuisance-erased probe is available, \conf{} selects it in every sample and $\cfn$ falls in every setting, model and sample (by \selFullConfCFmin--\selFullConfCFmax) at a mean \dec{} cost of at most \selFullConfDecMeanWorst.

\section{Discussion}

Our claim is narrower than ``probes are invalid'': decodability is not a reliable proxy for selectivity with respect to a named nuisance, and probes rely on a nuisance whenever their training data correlate it with the target. A claim that a direction \emph{measures} a concept should name plausible nuisances and report discriminant diagnostics and, where possible, nuisance-only edits for each; passing them is necessary evidence, not construct validity \citep{messick1995validity,licht2025measuring}.

\section*{Limitations}

Each setting examines one named nuisance (Amazon: two, separately); selectivity for unexamined nuisances is unknown. Edits are filtered by two LLM judges, not by new human annotation. We validate the judges against CEBaB's human labels and with planted errors, and recompute the results under every filter (Appendix~\ref{app:judgeval}), but human labels exist only for CEBaB, the judges agree only poorly to moderately with each other, and accepted pairs may be a selective subset; the code generates a blinded stratified sample for human annotation. LLM edits can change more than the nuisance: two LLM readers rate \rdWrongNeg{} of accepted negation edits (Amazon-Neg) as changing the opinion, its intensity or a fact, or as broken, mostly where a negation was added; negation edits can shift the intensity of an opinion despite the strength check. $\cfn$ is normalized by the class score gap, which is unstable for weak extractors, so we analyze competitive ones; an effect-size normalization (by the within-cell score SD, akin to Cohen's $d$) ranks extractors similarly ($\geq$\robSDMin). Correlated regimes are produced by resampling; they match natural rates for BiasBios ($P(\text{male}\mid\text{surgeon})=0.85$, $P(\text{female}\mid\text{nurse})=0.91$) and approximately CEBaB (0.80, 0.73) and Amazon-Neg, and are a stress test for Amazon-1P. Strict equivalence is established for few extractors on CEBaB. We study linear readouts of mean-pooled (and, as an ablation, last-token) representations of two 7--9B models; nonlinear probes, SAE features and other models are untested. Our measures concern what a score tracks in text, not what the model computes with it.

\section*{Ethics Statement}

BiasBios involves gender, and the appendix setting (Civil) religious and sexual identity. Concept scores used for auditing or monitoring can encode such attributes while passing accuracy checks; our diagnostics complement, and do not replace, fairness evaluation. The gender swap follows the dataset's binary annotation.

\bibliography{refs}

\appendix

\section{Edit validation and robustness}
\label{app:edits}
Table~\ref{tab:judges} reports, per judge and edit kind, how often the judge found the target kept, the nuisance changed, a fluency drop (edited text rated non-fluent while the source was rated fluent; both judges rate many untouched informal texts as non-fluent, so absolute fluency is not used), and a major other meaning change.\IfFileExists{tables/tab_agreement.tex}{ Table~\ref{tab:agree} gives the agreement of the two judges' accept decisions: agreement is low to moderate ($\kappa$ \kappaMin--\kappaMax), mostly because the Qwen judge is much stricter (it accepts \qwenPlaceboAccMin--\qwenPlaceboAccMax{} of placebo edits, gpt-oss-120b \gptPlaceboAccMin--\gptPlaceboAccMax), often reporting a fluency drop or a changed target for synonym swaps. Requiring both judges is therefore conservative, and acceptance by both is a filter, not verified correctness; rankings on accepted and on all edits agree closely (Table~\ref{tab:robust}). For Civil (Appendix~\ref{app:civil}), Qwen3-32B judges that swapping the named group changes toxicity in about half of the forward swaps, so few identity swaps survive.}{} The judges were never told which text was edited or which attribute was meant to change. The released code generates a blinded stratified sample of 25 pairs per dataset and edit kind for human annotation.

\paragraph{Amazon edits.} For each test review, gpt-oss-120b receives an instruction to add (or remove every instance of) the nuisance with an exact paraphrase that keeps every fact and the strength of the opinion, and returns JSON phrase replacements, which we apply by exact string match, so only the named phrases change; it may decline. Automatic checks require that the nuisance flips according to the word list, that at most 35\% of the words change, and that the length is stable. Examples of accepted edits: negation added, ``spaces between word were missing''$\to$``\dots{} not present''; removed, ``poorly written and not well thought out''$\to$``\dots{} ill-conceived'', ``I had not received the driver cd''$\to$``I had failed to receive \dots''; first person added, ``Received this book promptly''$\to$``I received \dots''; removed, ``is it just me?''$\to$``is this the only case?''. Prompts were developed on 100 test reviews per nuisance and frozen before a held-out check and the full run. The model declines most often when asked to add first person to a review without a natural place for it, so Amazon-1P has fewer additions than removals; every direction-specific result is reported separately (Table~\ref{tab:robust}). Because negation is part of how sentiment is expressed, a paraphrase that removes it can still sharpen an opinion; both judges answer an explicit question about the strength of the opinion, and the residual effect at $\rhot=0.5$ is discussed in \S\ref{sec:results}.

\begin{table*}[t]
\centering\small\setlength{\tabcolsep}{4pt}
\resizebox{0.85\textwidth}{!}{\begin{tabular}{@{}llrcccccc@{}}
\toprule
data & edit kind & $n$ & judge & $T$ kept & $N$ changed & fluency drop & major other & accepted \\
\midrule
Amazon-1P & nuisance & 1083 & Qwen3.8-27B & .98 & 1.00 & .21 & .00 & .74 \\
Amazon-1P & nuisance & 1083 & gpt-oss-120b & .98 & 1.00 & .16 & .00 & .82 \\
Amazon-1P & placebo & 567 & Qwen3.8-27B & .99 & .00 & .21 & .00 & .71 \\
Amazon-1P & placebo & 567 & gpt-oss-120b & .98 & .00 & .12 & .00 & .88 \\
Amazon-Neg & nuisance & 1427 & Qwen3.8-27B & .97 & 1.00 & .13 & .06 & .61 \\
Amazon-Neg & nuisance & 1427 & gpt-oss-120b & .98 & 1.00 & .03 & .02 & .90 \\
Amazon-Neg & placebo & 608 & Qwen3.8-27B & .99 & .00 & .21 & .00 & .72 \\
Amazon-Neg & placebo & 608 & gpt-oss-120b & .98 & .00 & .11 & .00 & .88 \\
BiasBios & nuisance & 1561 & Qwen3-32B & .98 & .99 & .03 & .01 & .94 \\
BiasBios & nuisance & 1561 & gpt-oss-120b & .99 & .99 & .01 & .00 & .96 \\
BiasBios & placebo & 961 & Qwen3-32B & .98 & .00 & .42 & .00 & .58 \\
BiasBios & placebo & 400 & gpt-oss-120b & .98 & .00 & .03 & .00 & .97 \\
CEBaB & nuisance & 210 & Qwen3-32B & .62 & .97 & .15 & .32 & .44 \\
CEBaB & nuisance & 210 & gpt-oss-120b & .93 & .98 & .05 & .22 & .69 \\
CEBaB & placebo & 158 & Qwen3-32B & .91 & .01 & .51 & .00 & .46 \\
CEBaB & placebo & 158 & gpt-oss-120b & 1.00 & .00 & .08 & .00 & .92 \\
CEBaB & target & 125 & Qwen3-32B & .00 & .08 & .10 & .27 & .55 \\
CEBaB & target & 125 & gpt-oss-120b & .02 & .10 & .03 & .16 & .76 \\
Civil & nuisance & 762 & Qwen3-32B & .45 & .98 & .13 & .72 & .17 \\
Civil & nuisance & 762 & gpt-oss-120b & 1.00 & .97 & .00 & .62 & .36 \\
Civil & nuisance rev & 400 & Qwen3-32B & .63 & .99 & .09 & .53 & .34 \\
Civil & nuisance rev & 400 & gpt-oss-120b & 1.00 & .98 & .00 & .36 & .62 \\
Civil & placebo & 480 & Qwen3-32B & .88 & .00 & .53 & .00 & .44 \\
Civil & placebo & 480 & gpt-oss-120b & 1.00 & .00 & .06 & .00 & .94 \\
Civil & placebo group & 400 & Qwen3-32B & .80 & .89 & .06 & .19 & .64 \\
Civil & placebo group & 400 & gpt-oss-120b & 1.00 & .90 & .00 & .06 & .94 \\
\bottomrule
\end{tabular}
}
\caption{LLM-judge verdicts per edit kind. Qwen judge: Qwen3-32B (Amazon: Qwen3.8-27B). Amazon: $N$ changed is the word-list check, and acceptance also requires an unchanged strength of opinion.}
\label{tab:judges}
\end{table*}

\IfFileExists{tables/tab_agreement.tex}{%
\begin{table}[h]
\centering\small\setlength{\tabcolsep}{3pt}
\resizebox{\columnwidth}{!}{\begin{tabular}{@{}llrcccc@{}}
\toprule
data & edit kind & $n$ & accept (Qwen3) & accept (gpt-oss) & both & $\kappa$ \\
\midrule
CEBaB & nuisance & 210 & .44 & .69 & .41 & .41 \\
CEBaB & placebo & 158 & .46 & .92 & .44 & .07 \\
CEBaB & target & 125 & .55 & .76 & .52 & .42 \\
BiasBios & nuisance & 1561 & .94 & .96 & .93 & .55 \\
BiasBios & placebo & 400 & .60 & .97 & .60 & .06 \\
Amazon-Neg & nuisance & 1427 & .61 & .90 & .58 & .18 \\
Amazon-Neg & placebo & 608 & .72 & .88 & .69 & .37 \\
Amazon-1P & nuisance & 1083 & .74 & .82 & .68 & .45 \\
Amazon-1P & placebo & 567 & .71 & .88 & .69 & .39 \\
Civil & nuisance & 762 & .17 & .36 & .13 & .31 \\
Civil & nuisance rev & 400 & .34 & .62 & .29 & .29 \\
Civil & placebo & 480 & .44 & .94 & .43 & .07 \\
Civil & placebo group & 400 & .64 & .94 & .62 & .08 \\
\bottomrule
\end{tabular}
}
\caption{Agreement of the two judges' accept decisions on pairs both rated.}
\label{tab:agree}
\end{table}}{}

\paragraph{Robustness of $\cfn$.} Table~\ref{tab:robust} gives rank correlations, across competitive extractors, between $\cfn$ computed on all vs.\ accepted edits; normalized by $\Delta_T$ vs.\ by the pooled within-cell score SD; signed vs.\ absolute; for the two edit directions separately (CEBaB, BiasBios, Amazon); and for the forward vs.\ reverse Civil swaps. At $\rhot=0.5$ on CEBaB the signed effect is small while the absolute change is not (median $|\cfn|$ \absCFcebabQFive{} vs.\ signed \sgnCFcebabQFive{}, Qwen), which is why the direction-wise rankings agree less there. Placebo edits are smaller than CEBaB's human rewrites, so the placebo floor is not magnitude-matched for CEBaB. \IfFileExists{tables/tab_placebo.tex}{\input{tables/tab_placebo}}{}

\begin{table}[h]
\centering\small\setlength{\tabcolsep}{2pt}
\resizebox{\columnwidth}{!}{\begin{tabular}{@{}lllccccc@{}}
\toprule
data & model & $\rhot$ & all vs.\ accepted & $\Delta_T$ vs.\ SD norm. & signed vs.\ $|\cdot|$ & direction $+$ vs.\ $-$ & Civil fwd vs.\ rev \\
\midrule
Amazon-1P & G & 0.5 & +.99 & 1.00 & +.69 & +.84 & -- \\
Amazon-1P & G & 0.7 & 1.00 & +.98 & +.91 & +.94 & -- \\
Amazon-1P & G & 0.9 & 1.00 & +.97 & +.99 & +.97 & -- \\
Amazon-1P & Q & 0.5 & +.99 & +.99 & +.46 & +.84 & -- \\
Amazon-1P & Q & 0.7 & 1.00 & +.97 & +.92 & +.94 & -- \\
Amazon-1P & Q & 0.9 & 1.00 & +.94 & 1.00 & +.98 & -- \\
Amazon-Neg & G & 0.5 & +.99 & +.98 & $-$.80 & +.78 & -- \\
Amazon-Neg & G & 0.7 & 1.00 & 1.00 & $-$.37 & +.86 & -- \\
Amazon-Neg & G & 0.9 & 1.00 & +.99 & +.93 & +.96 & -- \\
Amazon-Neg & Q & 0.5 & +.99 & +.98 & $-$.79 & +.79 & -- \\
Amazon-Neg & Q & 0.7 & +.99 & +.99 & $-$.21 & +.80 & -- \\
Amazon-Neg & Q & 0.9 & 1.00 & +.99 & +.92 & +.93 & -- \\
BiasBios & G & 0.5 & 1.00 & +.99 & $-$.94 & +.98 & -- \\
BiasBios & G & 0.7 & 1.00 & +.99 & +.95 & +.99 & -- \\
BiasBios & G & 0.9 & 1.00 & +.99 & 1.00 & 1.00 & -- \\
BiasBios & Q & 0.5 & 1.00 & +.99 & $-$.48 & +.97 & -- \\
BiasBios & Q & 0.7 & 1.00 & +.99 & +.97 & +.99 & -- \\
BiasBios & Q & 0.9 & 1.00 & +.98 & 1.00 & 1.00 & -- \\
CEBaB & G & 0.5 & +.77 & 1.00 & +.13 & +.31 & -- \\
CEBaB & G & 0.7 & +.93 & +.99 & +.89 & +.78 & -- \\
CEBaB & G & 0.9 & +.99 & +.97 & +.99 & +.94 & -- \\
CEBaB & Q & 0.5 & +.88 & +.99 & +.59 & +.69 & -- \\
CEBaB & Q & 0.7 & +.98 & +.99 & +.97 & +.94 & -- \\
CEBaB & Q & 0.9 & +.99 & +.87 & +.99 & +.94 & -- \\
Civil & G & 0.5 & +.98 & +.99 & $-$.24 & -- & +.82 \\
Civil & G & 0.7 & +.99 & +.99 & +.95 & -- & +.92 \\
Civil & G & 0.9 & +.98 & +.94 & +.99 & -- & +.76 \\
Civil & Q & 0.5 & +.93 & +.97 & +.42 & -- & +.76 \\
Civil & Q & 0.7 & +.89 & +.97 & +.99 & -- & +.68 \\
Civil & Q & 0.9 & +.92 & +.93 & 1.00 & -- & +.68 \\
\bottomrule
\end{tabular}
}
\caption{Rank correlations between variants of $\cfn$ across competitive extractors (medians of 3 training samples).}
\label{tab:robust}
\end{table}

\subsection{Validating the judges without new annotation}\label{app:judgeval}
\paragraph{Human labels (CEBaB).} CEBaB's edits were written by crowd workers, and five other workers labelled the food and service sentiment (positive, negative, not mentioned) of every original and edited review. We judged \jvN{} (original, food or service edit) pairs from the test and development families with both judges, including all \jvNbad{} edits whose intended aspect label did \emph{not} change according to the human majority (\jvBadRate{} of the set, oversampled). On whether each aspect is the same in the two texts, gpt-oss-120b agrees with the human majority in \jvGptQ{} of pairs, as often as a single held-out worker agrees with the majority of the other four (\jvHumanQ); Qwen3-32B agrees in \jvQwenQ, mostly because it reports changes the workers did not. Applying the acceptance rule, both judges reject \jvRejBad{} of the failed edits, and \jvPrec{} of accepted pairs are correct by the human labels (vs.\ \jvGoodRate{} of all judged pairs); the price is recall: half of the edits correct by the human labels are rejected (\jvRecall{} kept), in part for fluency or other meaning changes that the human labels do not cover.

\paragraph{Planted errors.} On edits both judges had accepted (50 per type and setting, disjoint sources), we planted one known error after the nuisance edit, keeping the word-list nuisance status: gpt-oss-120b proposed a phrase replacement that reverses the opinion, makes it clearly stronger or weaker, changes the occupation, or changes a minor fact; by rule, we garbled the text (swap, duplicate and drop a word) or reverted one gendered substitution (mixed gender cues). Table~\ref{tab:planted} gives how often the judges still accept. Reversed opinions, changed occupations and garbled text pass both judges in at most \plBadMax{} of cases; in the passing reversals the planted phrase mostly changes one aspect of a mixed review, not the overall opinion, and garbled reviews pass when the judges already rated the source non-fluent, which the rule allows. gpt-oss-120b alone accepts \plStrGptMin--\plStrGptMax{} of strength changes, so the strength check rests mainly on the Qwen judge (both: $\leq$\plStrBothMax). Mixed gender cues pass both in \plMixBoth. Minor fact changes pass often (\plFactMin--\plFactMax): the rule tolerates minor meaning changes by design, and some planted changes are trivial. Re-judging unmodified accepted edits in a new presentation order keeps \plCleanMin--\plCleanMax{} of them, a measure of judge noise.

\paragraph{Reading accepted Amazon edits.} Two further LLM readers not involved in editing or judging (Claude Opus 5.5 and GPT-6.1) each labelled the same 100 accepted Amazon edits (25 per setting and direction, judge verdicts hidden) as ok, unnatural (meaning kept, stilted wording), or wrong (opinion, strength or a fact changed, or broken text); they agree on \rdAgree{} of items ($\kappa=\rdKappa$). They rate \rdWrongFp{} of Amazon-1P edits wrong and \rdWrongNeg{} of Amazon-Neg edits, \rdWrongNegAdd{} of negation additions, which mostly replace a word by a weaker negated antonym (``one-of-a-kind''$\to$``not ordinary''); a further share is unnatural but meaning-preserving (any non-ok label: Amazon-1P \rdNotOkFp, Amazon-Neg \rdNotOkNeg). Removing first person from a superlative can broaden it (``the worst movie I have ever seen''$\to$``the worst movie ever seen''). Neither reader is a human, so this is a further check of edit quality, not a human validation.

\paragraph{Results under every filter.} Table~\ref{tab:filter} gives the near-best $\cfn$ range at $\rhot=0.9$ on all generated pairs, on pairs accepted by either judge alone, by at least one, and by both. Rankings of competitive extractors by $\cfn$ agree with those under the paper's filter (Spearman $\geq$\frSpearMin{} in every setting, $\rhot\in\{0.5,0.9\}$), and under every filter selecting by \conf{} with the erased probe lowers $\cfn$ in all eight settings at $\rhot=0.9$ (by \frSelFullMin--\frSelFullMax). Magnitudes depend on the filter only on CEBaB (\frCebabAll{} on all human edits vs.\ \frCebabBoth{} on accepted ones); rankings and the selection result hold there too.

\begin{table}[h]
\centering\small\setlength{\tabcolsep}{3pt}
\resizebox{\columnwidth}{!}{\begin{tabular}{@{}llccc@{}}
\toprule
data & planted error & Qwen & gpt-oss & both \\
\midrule
Amazon-Neg & none (retest) & .88 & .98 & .86 \\
 & opinion reversed & .08 & .14 & .06 \\
 & opinion stronger/weaker & .18 & .68 & .16 \\
 & garbled (rule) & .12 & .32 & .10 \\
 & minor fact changed & .76 & .88 & .70 \\
Amazon-1P & none (retest) & .98 & .92 & .92 \\
 & opinion reversed & .16 & .12 & .08 \\
 & opinion stronger/weaker & .18 & .58 & .18 \\
 & garbled (rule) & .16 & .26 & .10 \\
 & minor fact changed & .96 & .90 & .88 \\
BiasBios & none (retest) & .98 & .94 & .94 \\
 & occupation changed & .02 & .06 & .00 \\
 & gender cues mixed & .34 & .52 & .18 \\
 & garbled (rule) & .04 & .10 & .00 \\
 & minor fact changed & .94 & .52 & .52 \\
\bottomrule
\end{tabular}
}
\caption{Share of edits with a planted error that each judge accepts (50 per row). ``None (retest)'': unmodified accepted edits, re-judged.}
\label{tab:planted}
\end{table}

\begin{table*}[t]
\centering\small\setlength{\tabcolsep}{4pt}
\resizebox{\textwidth}{!}{\begin{tabular}{@{}lccccc@{}}
\toprule
 & all & Qwen & gpt-oss & either & both \\
\midrule
CEBaB & 210: .61--1.15/.41--.93 & 93: .47--.93/.26--.75 & 144: .56--1.01/.35--.81 & 151: .56--1.02/.35--.82 & 86: .46--.89/.26--.72 \\
BiasBios & 1561: .09--.41/.09--.33 & 1470: .09--.41/.09--.33 & 1498: .09--.41/.09--.33 & 1517: .09--.41/.09--.33 & 1451: .09--.42/.09--.33 \\
Amazon-Neg & 1427: .04--.13/.02--.13 & 870: .03--.12/.02--.12 & 1280: .04--.13/.02--.12 & 1316: .04--.13/.02--.12 & 834: .03--.12/.02--.12 \\
Amazon-1P & 1083: .11--.27/.06--.33 & 798: .11--.26/.06--.31 & 883: .11--.26/.06--.32 & 945: .11--.27/.06--.32 & 736: .11--.25/.05--.31 \\
\bottomrule
\end{tabular}
}
\caption{Near-best $\cfn$ at $\rhot=0.9$ (Qwen2.5-7B / Gemma-2-9B) under different edit filters; the number before the colon is the number of nuisance pairs.}
\label{tab:filter}
\end{table*}

\section{Further results}
\label{app:more}
\begin{figure*}[t]
\centering
\includegraphics[width=\textwidth]{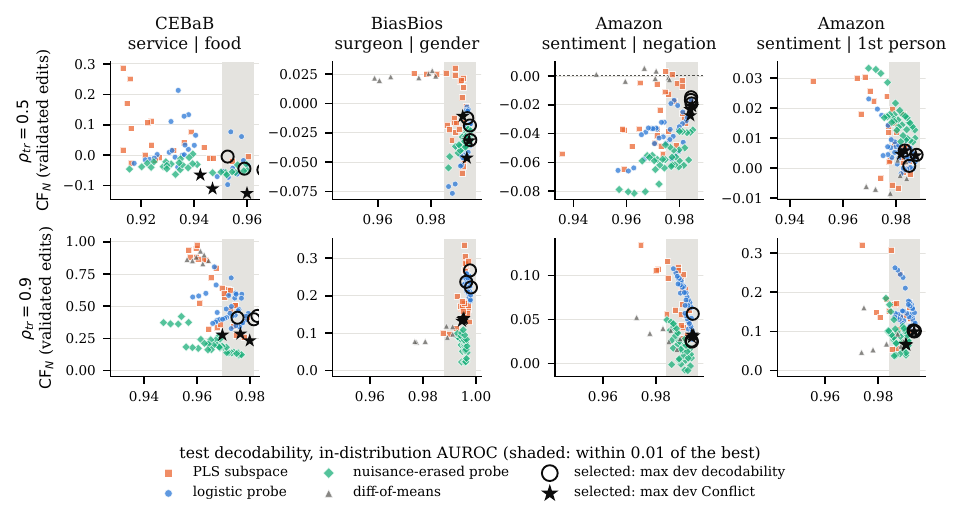}
\caption{As Figure~\ref{fig:main}, for Gemma-2-9B-it.}
\end{figure*}

\begin{table}[h]
\centering\small\setlength{\tabcolsep}{2pt}
\resizebox{\columnwidth}{!}{\begin{tabular}{@{}llccccc@{}}
\toprule
 & & best & \#near & $\cfn$ near-best & $\cfn$ equiv. & \conf{} \\
 & & \dec{} & (equiv.) & range & range & range \\
\midrule
\multicolumn{7}{@{}l}{\emph{$\rhot=0.7$}}\\
CEBaB & Q & .934 & 18 (0) & .15--.54 & -- & .70--.86 \\
 & G & .967 & 13 (0) & .05--.26 & -- & .86--.93 \\
BiasBios & Q & .995 & 62 (0) & .01--.14 & -- & .96--.99 \\
 & G & .995 & 61 (0) & $-$.01--.11 & -- & .96--.99 \\
Amazon-Neg & Q & .985 & 38 (0) & $-$.01--.03 & -- & .94--.96 \\
 & G & .988 & 48 (0) & $-$.01--.04 & -- & .94--.97 \\
Amazon-1P & Q & .988 & 27 (0) & .02--.08 & -- & .96--.98 \\
 & G & .989 & 56 (0) & .02--.09 & -- & .95--.98 \\
Civil & Q & .859 & 5 (0) & .14--.17 & -- & .64--.71 \\
 & G & .869 & 9 (0) & .09--.19 & -- & .66--.69 \\
\bottomrule
\end{tabular}
}
\caption{As Table~\ref{tab:near}, at $\rhot=0.7$ (equivalence tests were run at $\rhot\in\{0.5,0.9\}$ only).}
\end{table}

\begin{table*}[t]
\centering\small\setlength{\tabcolsep}{4pt}
\resizebox{0.8\textwidth}{!}{\begin{tabular}{@{}lllcccc@{}}
\toprule
data & model & $\rhot$ & $\cfn$ & $|\cfn|$ & \conf{} & $|$\leak{}$|$ \\
\midrule
Amazon-1P & G & 0.5 & $-$.43 [$-$.66, $-$.18] & $-$.66 [$-$.80, $-$.48] & +.96 [+.93, +.97] & +.47 [+.28, +.62] \\
Amazon-1P & G & 0.7 & $-$.31 [$-$.53, $-$.05] & $-$.51 [$-$.70, $-$.29] & +.94 [+.89, +.96] & $-$.11 [$-$.32, +.11] \\
Amazon-1P & G & 0.9 & $-$.00 [$-$.25, +.25] & $-$.06 [$-$.32, +.19] & +.68 [+.51, +.80] & +.32 [+.07, +.56] \\
Amazon-1P & Q & 0.5 & $-$.21 [$-$.45, +.05] & $-$.69 [$-$.79, $-$.50] & +.98 [+.95, +.99] & +.19 [$-$.08, +.43] \\
Amazon-1P & Q & 0.7 & $-$.58 [$-$.75, $-$.37] & $-$.71 [$-$.83, $-$.55] & +.94 [+.87, +.97] & $-$.20 [$-$.42, +.07] \\
Amazon-1P & Q & 0.9 & $-$.44 [$-$.64, $-$.18] & $-$.47 [$-$.69, $-$.23] & +.87 [+.80, +.91] & $-$.11 [$-$.36, +.16] \\
Amazon-Neg & G & 0.5 & +.26 [+.02, +.48] & $-$.74 [$-$.86, $-$.59] & +.95 [+.90, +.96] & +.09 [$-$.13, +.29] \\
Amazon-Neg & G & 0.7 & $-$.16 [$-$.37, +.07] & $-$.85 [$-$.92, $-$.76] & +.97 [+.93, +.98] & +.02 [$-$.22, +.27] \\
Amazon-Neg & G & 0.9 & $-$.40 [$-$.62, $-$.15] & $-$.50 [$-$.70, $-$.26] & +.73 [+.57, +.84] & $-$.13 [$-$.39, +.11] \\
Amazon-Neg & Q & 0.5 & +.25 [$-$.02, +.49] & $-$.68 [$-$.83, $-$.48] & +.94 [+.89, +.97] & +.01 [$-$.22, +.24] \\
Amazon-Neg & Q & 0.7 & $-$.31 [$-$.51, $-$.08] & $-$.87 [$-$.93, $-$.76] & +.96 [+.93, +.97] & $-$.17 [$-$.39, +.05] \\
Amazon-Neg & Q & 0.9 & $-$.26 [$-$.50, +.00] & $-$.41 [$-$.63, $-$.18] & +.79 [+.68, +.86] & $-$.06 [$-$.31, +.22] \\
BiasBios & G & 0.5 & $-$.33 [$-$.54, $-$.07] & +.06 [$-$.18, +.32] & +.90 [+.82, +.94] & $-$.24 [$-$.45, $-$.04] \\
BiasBios & G & 0.7 & +.41 [+.22, +.55] & +.41 [+.22, +.55] & +.69 [+.50, +.83] & +.59 [+.44, +.71] \\
BiasBios & G & 0.9 & +.38 [+.09, +.60] & +.38 [+.11, +.61] & +.52 [+.33, +.70] & +.40 [+.14, +.61] \\
BiasBios & Q & 0.5 & $-$.19 [$-$.43, +.08] & $-$.50 [$-$.67, $-$.27] & +.91 [+.84, +.95] & $-$.16 [$-$.38, +.10] \\
BiasBios & Q & 0.7 & +.05 [$-$.20, +.29] & +.05 [$-$.18, +.29] & +.75 [+.59, +.85] & +.35 [+.12, +.54] \\
BiasBios & Q & 0.9 & +.11 [$-$.16, +.38] & +.11 [$-$.17, +.37] & +.53 [+.31, +.70] & +.32 [+.06, +.54] \\
CEBaB & G & 0.5 & $-$.35 [$-$.58, $-$.05] & $-$.71 [$-$.83, $-$.51] & +.80 [+.65, +.89] & $-$.19 [$-$.45, +.10] \\
CEBaB & G & 0.7 & $-$.43 [$-$.65, $-$.17] & $-$.61 [$-$.77, $-$.37] & +.66 [+.43, +.80] & $-$.29 [$-$.54, +.01] \\
CEBaB & G & 0.9 & $-$.61 [$-$.75, $-$.44] & $-$.65 [$-$.78, $-$.46] & +.68 [+.50, +.80] & $-$.47 [$-$.62, $-$.24] \\
CEBaB & Q & 0.5 & $-$.23 [$-$.50, +.09] & $-$.65 [$-$.83, $-$.39] & +.74 [+.57, +.86] & $-$.08 [$-$.37, +.24] \\
CEBaB & Q & 0.7 & $-$.50 [$-$.69, $-$.24] & $-$.58 [$-$.73, $-$.37] & +.69 [+.53, +.79] & $-$.37 [$-$.57, $-$.12] \\
CEBaB & Q & 0.9 & $-$.43 [$-$.59, $-$.22] & $-$.46 [$-$.62, $-$.26] & +.35 [+.16, +.53] & +.20 [+.01, +.39] \\
Civil & G & 0.5 & +.05 [$-$.30, +.38] & $-$.51 [$-$.70, $-$.25] & +.67 [+.39, +.85] & +.43 [+.11, +.67] \\
Civil & G & 0.7 & $-$.20 [$-$.50, +.16] & $-$.30 [$-$.59, +.04] & +.24 [$-$.10, +.53] & +.53 [+.16, +.82] \\
Civil & G & 0.9 & +.03 [$-$.21, +.26] & $-$.02 [$-$.26, +.22] & $-$.25 [$-$.51, +.02] & +.66 [+.41, +.83] \\
Civil & Q & 0.5 & $-$.09 [$-$.41, +.26] & $-$.42 [$-$.63, $-$.11] & +.70 [+.43, +.83] & +.17 [$-$.23, +.50] \\
Civil & Q & 0.7 & $-$.01 [$-$.32, +.32] & $-$.08 [$-$.40, +.26] & +.24 [$-$.11, +.56] & +.53 [+.19, +.80] \\
Civil & Q & 0.9 & $-$.01 [$-$.29, +.30] & $-$.00 [$-$.30, +.29] & $-$.47 [$-$.68, $-$.21] & +.74 [+.47, +.91] \\
\bottomrule
\end{tabular}
}
\caption{Spearman correlation of test \dec{} with each facet across competitive standard extractors, with 95\% bootstrap CIs over extractors. Hyperparameter variants of one family are not independent, so CIs are descriptive.}
\end{table*}

\begin{table}[h]
\centering\small\setlength{\tabcolsep}{2pt}
\resizebox{\columnwidth}{!}{\begin{tabular}{@{}lllcccc@{}}
\toprule
data & model & $\rhot$ & $\Delta$\dec{} [90\% CI] & equiv. & $\cfn$ probe & $\cfn$ erased \\
\midrule
Amazon-1P & G & 0.5 & +.000 [$-$.000, +.000] & yes & .00 & .01 \\
Amazon-1P & G & 0.9 & +.003 [+.001, +.005] & yes & .10 & .04 \\
Amazon-1P & Q & 0.5 & +.000 [$-$.000, +.000] & yes & .00 & .01 \\
Amazon-1P & Q & 0.9 & +.002 [+.000, +.004] & yes & .17 & .07 \\
Amazon-Neg & G & 0.5 & $-$.001 [$-$.001, $-$.000] & yes & $-$.01 & $-$.04 \\
Amazon-Neg & G & 0.9 & +.001 [+.000, +.001] & yes & .03 & $-$.00 \\
Amazon-Neg & Q & 0.5 & $-$.001 [$-$.001, $-$.000] & yes & $-$.03 & $-$.05 \\
Amazon-Neg & Q & 0.9 & +.001 [+.001, +.002] & yes & .04 & $-$.00 \\
BiasBios & G & 0.5 & $-$.000 [$-$.000, +.000] & yes & $-$.01 & $-$.03 \\
BiasBios & G & 0.9 & +.001 [$-$.000, +.002] & yes & .24 & .07 \\
BiasBios & Q & 0.5 & +.000 [$-$.000, +.000] & yes & $-$.00 & $-$.02 \\
BiasBios & Q & 0.9 & +.001 [+.000, +.002] & yes & .22 & .07 \\
CEBaB & G & 0.5 & +.000 [$-$.000, +.001] & yes & $-$.00 & $-$.01 \\
CEBaB & G & 0.9 & +.004 [$-$.003, +.012] & no & .63 & .14 \\
CEBaB & Q & 0.5 & $-$.003 [$-$.006, +.000] & yes & .13 & .02 \\
CEBaB & Q & 0.9 & +.009 [$-$.003, +.023] & no & .83 & .34 \\
Civil & G & 0.5 & +.008 [+.005, +.011] & no & .01 & $-$.01 \\
Civil & G & 0.9 & +.039 [+.030, +.049] & no & .26 & .18 \\
Civil & Q & 0.5 & +.011 [+.006, +.015] & no & .06 & .04 \\
Civil & Q & 0.9 & +.049 [+.035, +.063] & no & .40 & .28 \\
\bottomrule
\end{tabular}
}
\caption{Logistic probe vs.\ nuisance-erased probe, each family's layer and hyperparameter chosen on development \dec{} (training sample 0). Equiv.: 90\% CI of the \dec{} difference within $\pm0.01$.}
\end{table}

\paragraph{Readout ablation.} Reading the last non-special token instead of mean pooling (Qwen2.5-7B, Table~\ref{tab:pooling}) preserves the pattern: under correlated training, near-best extractors differ substantially in $\cfn$, decorrelated training largely removes it, and Conflict-based selection lowers $\cfn$. On Amazon-Neg there is little to lower under either readout. On Civil the last-token readout is less decodable and the selection costs considerably more accuracy.

\IfFileExists{tables/tab_pooling.tex}{\begin{table}[h]
\centering\small\setlength{\tabcolsep}{2pt}
\resizebox{\columnwidth}{!}{%
\begin{tabular}{@{}llccccc@{}}
\toprule
 & & \multicolumn{2}{c}{$\rhot=0.9$} & \multicolumn{2}{c}{$\rhot=0.5$} & selection, $\rhot=0.9$ \\
data & readout & best \dec{} & $\cfn$ near-best & best \dec{} & $\cfn$ near-best & $\cfn$: \dec{}$\to$\conf{} ($\Delta$\dec{}) \\
\midrule
CEBaB & mean & .966 & .46--.89 & .924 & .02--.28 & .69$\to$.54 ($-$.012) \\
CEBaB & last token & .968 & .51--.76 & .921 & .15--.27 & .56$\to$.49 ($-$.003) \\
BiasBios & mean & .998 & .09--.42 & .995 & $-$.04--.05 & .22$\to$.13 ($-$.001) \\
BiasBios & last token & .998 & .10--.74 & .994 & $-$.06--.01 & .31$\to$.11 ($-$.005) \\
Amazon-Neg & mean & .992 & .03--.12 & .981 & $-$.06--$-$.00 & .04$\to$.04 (+.000) \\
Amazon-Neg & last token & .990 & .01--.03 & .977 & $-$.01--.01 & .02$\to$.02 ($-$.000) \\
Amazon-1P & mean & .993 & .11--.25 & .986 & $-$.01--.03 & .17$\to$.11 ($-$.002) \\
Amazon-1P & last token & .987 & .05--.13 & .979 & $-$.01--.03 & .09$\to$.06 ($-$.001) \\
Civil & mean & .932 & .24--.45 & .829 & .04--.08 & .43$\to$.33 ($-$.018) \\
Civil & last token & .904 & .67--.67 & .794 & .01--.05 & .67$\to$.09 ($-$.156) \\
\bottomrule
\end{tabular}}
\caption{Readout ablation (Qwen2.5-7B, standard extractors, accepted edits): mean pooling vs.\ the last non-special token.}
\label{tab:pooling}
\end{table}
}{}

\begin{table*}[t]
\centering\small
\resizebox{\textwidth}{!}{\begin{tabular}{@{}ll|cccc|ccc@{}}
\toprule
 & & \multicolumn{4}{c|}{standard extractors: $\cfn$ ($\Delta$\dec{})} & \multicolumn{3}{c}{+ erased probe: $\cfn$ ($\Delta$\dec{})} \\
 & & max \dec{} & max \conf{} & min $|$\leak{}$|$ & \dec{}$-|$\leak{}$|$ & max \conf{} & min $|$\leak{}$|$ & \dec{}$-|$\leak{}$|$ \\
\midrule
CEBaB & Q & .69 & .54 ($-$.012) & .50 ($-$.013) & .52 ($-$.008) & .34 ($-$.007) & .34 ($-$.007) & .33 ($-$.011) \\
 & G & .41 & .26 ($-$.005) & .27 ($-$.004) & .26 ($-$.005) & .11 ($-$.004) & .12 ($-$.004) & .14 ($-$.010) \\
BiasBios & Q & .22 & .13 ($-$.001) & .09 ($-$.004) & .11 ($-$.021) & .03 ($-$.002) & .03 ($-$.002) & .03 ($-$.002) \\
 & G & .24 & .14 ($-$.002) & .12 ($-$.006) & .08 ($-$.017) & .02 ($-$.002) & .02 ($-$.001) & .02 ($-$.001) \\
Amazon-Neg & Q & .04 & .04$^\dagger$ (+.000) & .07$^\dagger$ ($-$.007) & .05$^\dagger$ ($-$.009) & $-$.00 ($-$.000) & $-$.00 ($-$.000) & $-$.00 ($-$.000) \\
 & G & .04 & .03$^\dagger$ (+.000) & .06$^\dagger$ ($-$.004) & .02$^\dagger$ ($-$.002) & $-$.00 ($-$.001) & $-$.01 ($-$.002) & $-$.01 ($-$.002) \\
Amazon-1P & Q & .17 & .11 ($-$.002) & .11 ($-$.002) & .11 ($-$.002) & .07 ($-$.002) & .06 ($-$.008) & .06 ($-$.013) \\
 & G & .10 & .08$^\dagger$ ($-$.002) & .08$^\dagger$ ($-$.007) & .08$^\dagger$ ($-$.007) & .04 ($-$.003) & .04 ($-$.004) & .04 ($-$.004) \\
Civil & Q & .43 & .33 ($-$.018) & .31 ($-$.001) & .37 ($-$.041) & .15 ($-$.034) & .31 ($-$.001) & .17 ($-$.038) \\
 & G & .31 & .20 ($-$.015) & .30$^\dagger$ ($-$.006) & .22$^\dagger$ ($-$.020) & .11 ($-$.036) & .30$^\dagger$ ($-$.006) & .11 ($-$.034) \\
\bottomrule
\end{tabular}
}
\caption{All development-set selection criteria at $\rhot=0.9$ (as Table~\ref{tab:select}), for standard extractors and with the nuisance-erased probe.}
\end{table*}

\section{Civil Comments: rule-based identity swaps}
\label{app:civil}
\textbf{CivilComments} (WILDS; \citealp{borkan2019nuanced,koh2021wilds}) pairs toxicity ($T$) with a human annotation that the comment mentions Muslims or LGBTQ people ($N$). Following \citet{dixon2018measuring} and \citet{garg2019counterfactual}, we swapped those identity terms for Christian/straight terms and, on 400 further comments, ``Christian'' for ``Muslim'', averaging the two directions; placebos also include ``Christian''$\to$``Protestant''. Both judges reject most of these swaps (Table~\ref{tab:data}: \civilAccFwd{} and \civilAccRev{} accepted by both), mostly because swapping the group changes a claim about it (``the baker discriminated against the gay couple''), not just its name. We therefore report Civil here only. Its results agree with the main settings: at $\rhot=0.9$ near-best extractors have $\cfn$ \nbCFmincivilQNine--\nbCFmaxcivilQNine{} (Qwen; Gemma \nbCFmincivilGNine--\nbCFmaxcivilGNine) and $\conf{}<0.5$ throughout, i.e.\ the most decodable toxicity scores rank non-toxic comments about Muslims or LGBTQ people above toxic comments that mention neither; at $\rhot=0.5$ the net shift is small (\nbCFmincivilQFive--\nbCFmaxcivilQFive); and choosing by \conf{} lowers $\cfn$ at a larger accuracy cost than in the main settings, because the nuisance is predictive in-distribution (Table~\ref{tab:near-all} and the selection table in Appendix~\ref{app:more}).

\begin{table}[h]
\centering\small\setlength{\tabcolsep}{2pt}
\resizebox{\columnwidth}{!}{\begin{tabular}{@{}llccccc@{}}
\toprule
 & & best & \#near & $\cfn$ near-best & $\cfn$ equiv. & \conf{} \\
 & & \dec{} & (equiv.) & range & range & range \\
\midrule
\multicolumn{7}{@{}l}{\emph{$\rhot=0.9$ (correlated training data)}}\\
CEBaB & Q & .966 & 37 (9) & .46--.89 & .54--.83 & .44--.69 \\
 & G & .979 & 39 (8) & .26--.72 & .25--.63 & .57--.82 \\
BiasBios & Q & .998 & 66 (65) & .09--.42 & .09--.42 & .83--.97 \\
 & G & .998 & 66 (61) & .09--.33 & .11--.33 & .92--.97 \\
Amazon-Neg & Q & .992 & 51 (45) & .03--.12 & .03--.12 & .87--.95 \\
 & G & .994 & 64 (60) & .02--.12 & .02--.11 & .87--.96 \\
Amazon-1P & Q & .993 & 40 (27) & .11--.25 & .11--.22 & .85--.95 \\
 & G & .994 & 62 (49) & .05--.31 & .06--.25 & .84--.96 \\
Civil & Q & .932 & 8 (5) & .24--.45 & .28--.42 & .35--.41 \\
 & G & .940 & 12 (1) & .30--.50 & .35 & .29--.43 \\
\midrule
\multicolumn{7}{@{}l}{\emph{$\rhot=0.5$ (decorrelated)}}\\
CEBaB & Q & .924 & 10 (1) & .02--.28 & .09 & .84--.90 \\
 & G & .960 & 11 (1) & $-$.10--.08 & .02 & .93--.95 \\
BiasBios & Q & .995 & 58 (54) & $-$.04--.05 & $-$.04--.05 & .98--.99 \\
 & G & .995 & 60 (53) & $-$.08--.03 & $-$.06--.03 & .98--.99 \\
Amazon-Neg & Q & .981 & 27 (14) & $-$.06--$-$.00 & $-$.05--$-$.01 & .95--.97 \\
 & G & .985 & 34 (23) & $-$.06--.00 & $-$.05--.00 & .95--.97 \\
Amazon-1P & Q & .986 & 28 (16) & $-$.01--.03 & $-$.01--.01 & .97--.98 \\
 & G & .987 & 53 (34) & $-$.01--.02 & $-$.01--.01 & .97--.99 \\
Civil & Q & .829 & 4 (2) & .04--.08 & $-$.00--.03 & .81--.83 \\
 & G & .839 & 5 (3) & $-$.02--.07 & $-$.03--.03 & .83--.84 \\
\bottomrule
\end{tabular}
}
\caption{As Table~\ref{tab:near}, all settings including Civil (accepted edits; Civil: mean of both swap directions).}
\label{tab:near-all}
\end{table}

\section{Synthetic representation}
\label{app:synthetic}
\begin{figure}[h]
\centering
\includegraphics[width=\columnwidth]{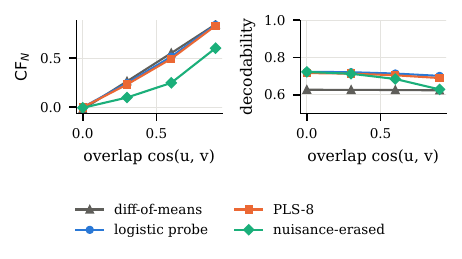}
\caption{Synthetic representation, decorrelated training ($\rhot=0.5$): nuisance sensitivity (left) and decodability (right) as target--nuisance overlap grows; means over 5 replicates.}
\label{fig:synth}
\end{figure}

We generate $x=T\,u+N\,v+\varepsilon$ in $\mathbb{R}^{256}$ with unit vectors $u,v$ at a chosen cosine (``overlap''), binary $T$ and $N$ with training agreement $\rhot$, and anisotropic Gaussian noise (power-law spectrum). A nuisance edit is exact: $x'=x\pm v$. We fit the same extractors with the same code and compute $\cfn$ on a decorrelated pool. With no overlap, $\cfn$ grows with training correlation (from $\approx0$ at $\rhot=0.5$ to $\approx0.75$ at $0.9$) for every standard extractor: the familiar shortcut. With decorrelated training, $\cfn$ grows with overlap (by \synCFRangeStdMin--\synCFRangeStdMax{} across overlaps for standard extractors) while decodability changes relatively little (by at most \synDecRangeStd; \synDecRangeNe{} for the erased probe; Figure~\ref{fig:synth}). In this low signal-to-noise setting even logistic probes inherit the overlap; in CEBaB, where the target is decoded far better, they inherit much less of it (\S\ref{sec:results}). The synthetic setting illustrates that decodability is largely insensitive to overlap; it is not a model of the real representations.

\section{Reproducibility and implementation audit}
\label{app:repro}
Code, data construction, judge outputs and per-extractor metrics are in the repository linked on the first page; every table and in-text number in this paper is generated from saved results by a script. An automated audit (released) checks that: training, development and test texts are disjoint (CEBaB at the level of review families; two BiasBios test texts and one development text that duplicated training texts were removed); every edit's orientation matches its labels; PLS with $k{=}1$ reproduces the difference of means numerically (cosine 1.000, same orientation); no evaluated text is truncated at 256 tokens on CEBaB, BiasBios and Amazon, and 0.2\% on Civil; the Amazon training, development and test pools share no text; and an independent re-implementation of every metric from the saved raw scores (rank-based AUROC, no shared code) matches the pipeline to $<10^{-8}$. Layers: Qwen hidden states 7, 10, 14, 18, 21, 24, 28 of 28; Gemma 10, 16, 21, 26, 32, 37, 42 of 42. Bootstrap CIs resample review families (CEBaB) or texts within $(T,N)$ cells, and edit pairs.

We do not report residual ``completeness'' \citep{elazar2021amnesic}: whether linear information survives projecting out a subspace is largely fixed by whether the subspace contains the class-mean difference \citep{belrose2023leace}, a property of the estimator rather than of what its score measures.

\end{document}